\documentclass{article}

\usepackage{arxiv}

\usepackage[utf8]{inputenc} % allow utf-8 input
\usepackage[T1]{fontenc}    % use 8-bit T1 fonts
\usepackage[colorlinks=true,citecolor=blue,urlcolor=blue,linkcolor=blue]{hyperref}       % hyperlinks
\usepackage{url}            % simple URL typesetting
\usepackage{booktabs}       % professional-quality tables
\usepackage{amsfonts}       % blackboard math symbols
\usepackage{nicefrac}       % compact symbols for 1/2, etc.
\usepackage{microtype}      % microtypography
\usepackage{soul, color}
\usepackage{wrapfig}
\usepackage{charter}
\usepackage{amsmath}
\usepackage{graphicx}
\usepackage{setspace}
\usepackage{multicol}
\usepackage{stfloats}
\usepackage[hang,flushmargin]{footmisc}
\usepackage[sorting=none,maxbibnames=50,urldate=iso,date=iso,giveninits=true,seconds=true,doi=false,isbn=false,eprint=false]{biblatex}
\title{Artificial Intelligence Narratives: An Objective Perspective on Current Developments}

\author{
  Noah Klarmann\thanks{primary contact: noah.klarmann@tum.de} \\
  Chair of Robotics, AI and Real-time Systems\\
  Department of Informatics\\
  Technical University of Munich\\
}

\begin{document}
%\onehalfspacing
%\doublespacing

\maketitle

\begin{abstract}
This work provides a starting point for researchers interested in gaining a deeper understanding of the big picture of artificial intelligence (AI).
To this end, a narrative is conveyed that allows the reader to develop an objective view on current developments that is free from false promises that dominate public communication.
An essential takeaway for the reader is that AI must be understood as an umbrella term encompassing a plethora of different methods, schools of thought, and their respective historical movements.
Consequently, a bottom-up strategy is pursued in which the field of AI is introduced by presenting various aspects that are characteristic of the subject.
This paper is structured in three parts: 
(i) Discussion of current trends revealing false public narratives, 
(ii) an introduction to the history of AI focusing on recurring patterns and main characteristics, and 
(iii) a critical discussion on the limitations of current methods in the context of the potential emergence of a strong(er) AI.
It should be noted that this work does not cover any of these aspects holistically; rather, the content addressed is a selection made by the author and subject to a didactic strategy.
\end{abstract}
%
%\keywords{Current AI Developments, Overview to Artifical Intelligence, AGI, AI History, AI Milestones}
%
\begin{multicols}{2}
\section{Today's Perspectives on AI}
Undoubtedly, advances in AI have enormous implications on societies, politics, and industries, which has resulted in an increasing attention over the past few years.
The tremendous value from leveraging AI-based approaches is becoming more and more apparent in a wide variety of different areas, such as 
(i) drug discovery in medicine (an example is given by \textcite{Ramsundar.07022015}), 
(ii) theoretical research where AI allows to systematically search hypothesis space to find unbiased, repeatable, and optimal results based on big data sets (for instance, see \cite{Gil.2014}), 
or (iii) transportation where researchers strive for fully autonomous driving \cite{Levinson.2011}.
Especially data-driven technologies are already transforming entire industries by improving processes or paving the way for fundamentally new business models \cite{Brynjolfsson.2014}.
It's worth noting that the huge popularity of AI is fueled by the fact that the world's most successful companies sit on enormous data lakes of mostly personal data, forming powerful monopolies changing the way we live (\textit{Google}\texttrademark, \textit{Facebook}\texttrademark, or \textit{Amazon}\texttrademark).
Commercially viable business models are typically built on exceedingly narrow models to be competitive as - similar to humans - specialization usually leads to superior performance.
Products from big tech corporates are ubiquitous in our everyday lives and may have led to a widespread misunderstanding of what AI is.
Truth to be told, it is exceptionally difficult to define the term \textit{intelligence} on its own, which naturally also applies to the term (AI).
To complicate comprehension of the term "AI" even further, the societal view of AI is taken ad absurdum by unscientific debates or science fiction \cite{Lorencik.}.
In addition, one can often observe unrealistic product advertisement when new products based on ordinary software development are sold as being based on advanced AI techniques, as "AI" is currently a very marketable keyword.
Besides the confusion that is conveyed by public communication, the field of AI is difficult to understand as it comprises a wide variety of different schools of thought and methodologies.
Furthermore, the public discussion is dominated by extremes.
On the one hand, AI evangelists appear who push away any concerns.
On the other hand, there are people who advocate an overly skeptical view on AI; for example, by portraying a dystopian world in which robots enslave humanity.
It is obvious that all these mechanisms inevitably have an enormous impact on the way AI is viewed today.
For instance, young people of generations "Y" and "Z" mostly associate AI with business models from big tech companies that exploit AI, as this is the most visible aspect of their lifetime to date.
At the same time, individuals from the "Boomer" and "X" generations are possibly more skeptical about current developments in AI, primarily due to two distinct aspects that characterized the field of AI in the 20th century: 
(i) Reaching milestones usually caused disappointment when the anticipated technological revolution did not occur.
For example, IBM's Deep Blue defeated the world grandmaster Garri Kasparov in 1997 \cite{Campbell.2002} that lead to great attention in the scientific world and also beyond.
Apparently, it was a great surprise to the public when the agent that is able to master such an intellectual task as chess would not made a good lawyer or taxi driver.  
(ii) It has happened several times that entire industries with questionable AI-based business models have perished when high expectations led to disappointment among society and stakeholders as extravagant promises remained unfulfilled.
Although skepticism and misconceptions are evident, AI is undoubtedly playing an increasingly important role in our daily lives, leading to growing concerns about privacy \cite{Ji.24122014}, security \cite{Papernot.2017}, fairness \cite{Johnson.01082016}, or the threat posed by autonomous military agents \cite{AIRoboticsresearchers4502andothers.2015}.
Consequently, regulation and standardization of AI methods are crucial to constrain the development in a way that corresponds to mankind's expectations \cite{Tegmark.2017,Bostrom.2016}.
In this context, two concerns commonly appear:
(i) Higher degrees of automation are associated with job losses as machines substitute human labor.
In fact, a survey among researchers revealed a predicted chance of 50\% that AI will be superior to human in almost every task within the next 42 years (until 2063) and could replace all human labor within the next 117 years (by 2138) \cite{Grace.2018}.
However, it should be noted that substituted jobs do not necessarily lead to unemployment in the short term.
Miller and Atkinson \cite{B.Miller.2013} advocate that the "job loss" argument is based on a common fallacy, namely the claim that the number of jobs decreases with increasing automation/productivity.
In history however, no correlation between productivity and unemployment can be identified.
The wrong assumption of job losses is based on the hypothesis that there is a limit amount of labor that does not rise with increasing productivity.
This is a wrong assumption as savings due to increased productivity recycles back to economy, rising the demand for products that in turn lead to more jobs.
In addition, new jobs are also created in the prospering automation/AI industry \cite{B.Miller.2013}.
It is worth noting that a major risk for industrial ecosystems is to refuse the highest possible level of automation, as this would lead to a loss of competitiveness.
(ii) A second often raised issue is that advanced techniques are black-box models and hence, cannot be controlled and might sooner or later emerge towards something with an own will (see for instance \cite{Tegmark.2017}).
While it is true that many of the currently used methods are black-box models (for example, machine learning), they are by no means self-organizing or emergent (yet).
It is also worth noting that significant efforts are being made by researchers to develop explainable AI to make use of machine learning with more caution (especially in safety-critical areas such as autonomous driving or in the medical field) or to derive knowledge from machine learning models (an overview on this topic can be found by \textcite{Tjoa.2020}).

Despite the great success in employing AI methods, there is a significant lack of understanding in society, as well as among scientists.
To this end, this paper provides an objective overview of the field of AI by addressing the most important characteristics.
The author is convinced that this bottom-up approach (starting from what AI encompasses) is the best strategy from a didactic point of view. 
The alternative (a top-down approach) would imply to introduce a high-level definition that has to be exceedingly broad/complex and would hence be difficult to formulate/understand.
In the next section, an introduction to the history of AI is given. 
Different eras that were prevalent in the past as well as important events, such as Milestones, are presented.
A look at the recent past shows that trends in AI are often temporary, may change significantly over time, were adopted by other methods, or branch into different disciplines.
In the subsequent section, an outlook on possible future developments is given.

%\newpage
\section{Characteristics of AI history}
%
%\begin{wrapfigure}{r}{0.52\textwidth}
    %\centering%
    %\includegraphics[width=.5\textwidth]{figures/1_ai_by_meth.pdf}%
    %\caption{Historical epochs of AI (extended and adapted from \cite{Russell.2016, Launchbury.})}%
    %\label{1_ai_by_meth}%
%\end{wrapfigure}%
%
%\end{multicols}

\begin{figure*}%{r}{0.52\textwidth}
    \centering%
    \includegraphics[width=.95\textwidth]{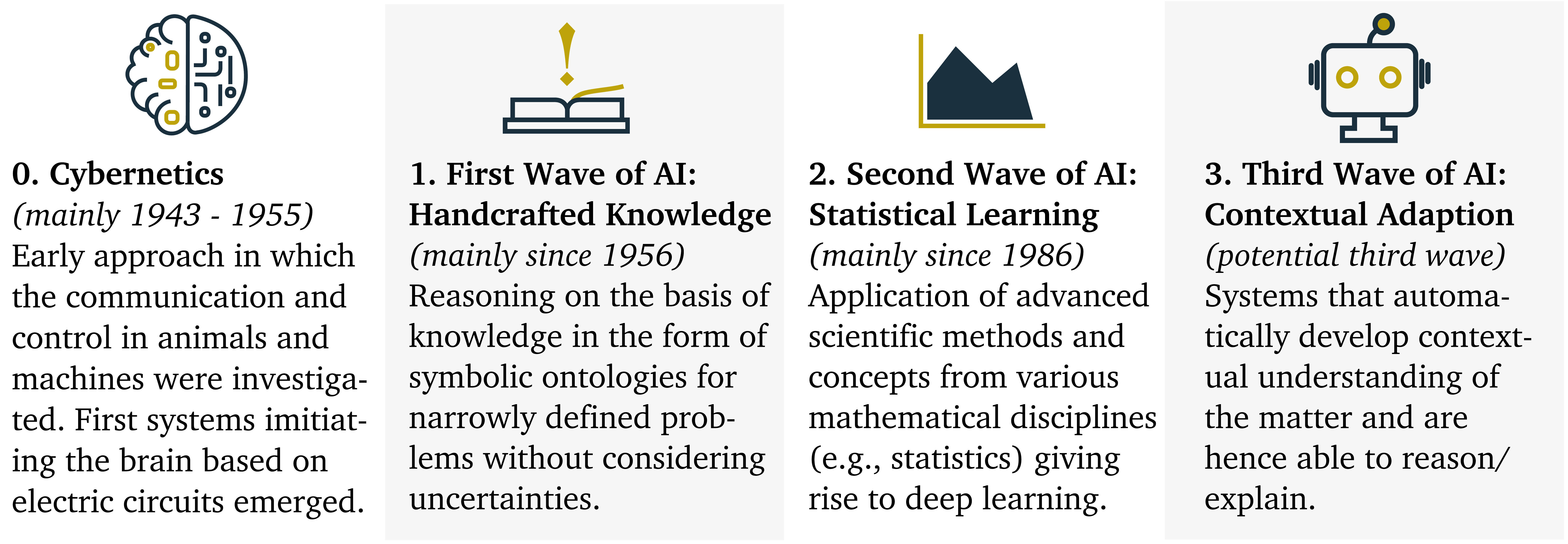}%
    \caption{Historical epochs of AI (extended and adapted from \textcite{Russell.2016, Launchbury.})}%
    \label{1_ai_by_meth}%
\end{figure*}%
%\begin{multicols}{2}}

\begin{figure*} %sidewaysfigure
    \centering%
    \includegraphics[width=1.\textwidth]{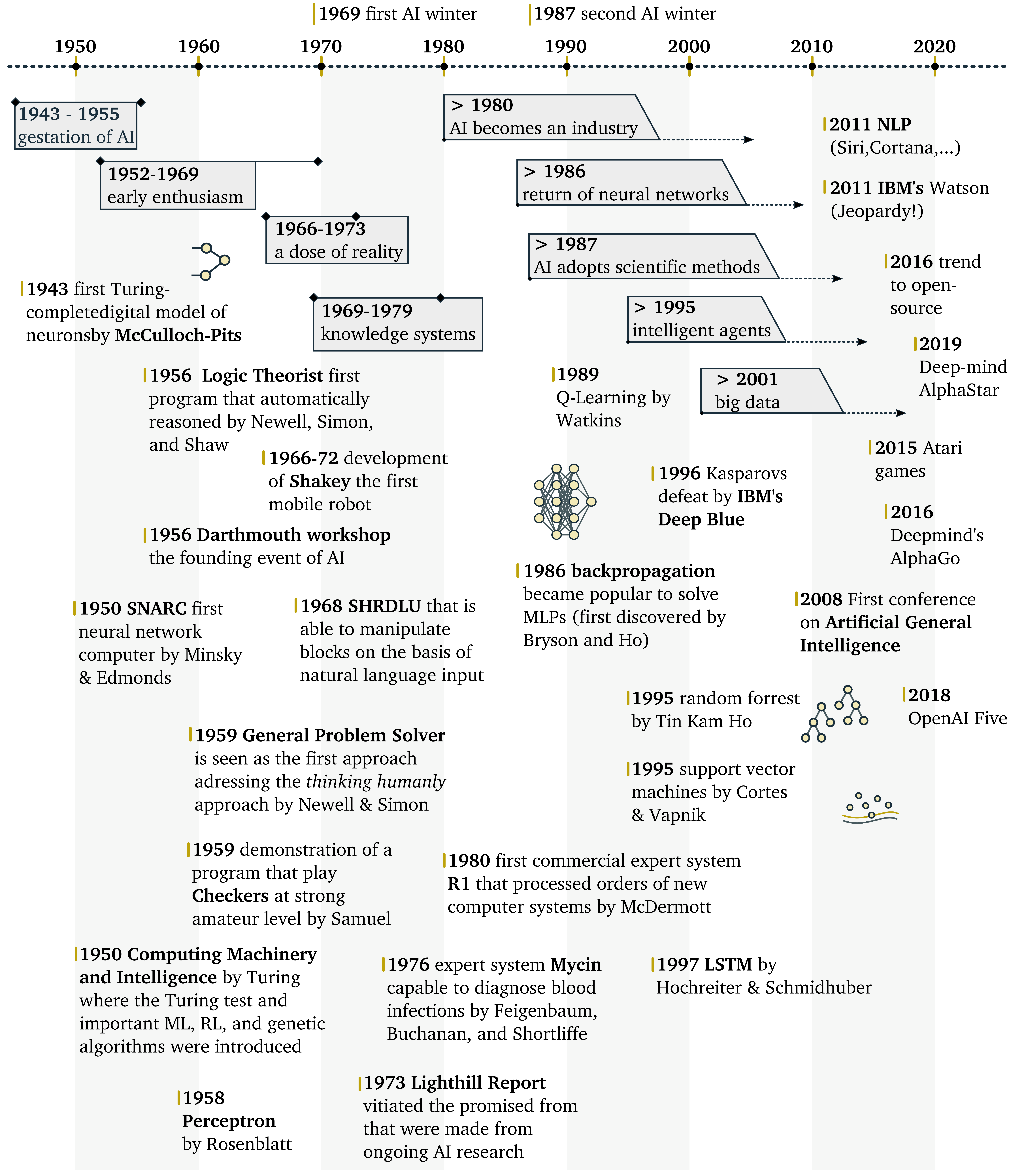}%
    \caption{Most important phases and milestones in the field of AI (mostly inspired by \textcite{Russell.2016, Launchbury.}).}%
    \label{1_timeline}%
\end{figure*}

A brief introduction to the history of AI is provided in this section.
Besides conveying known facts, main characteristics of AI are presented to support the reader in the comprehension of current developments and to derive possible scenarios for the future.
On a high level, the history of AI can be categorized into four phases, as shown in Figure \ref{1_ai_by_meth}.
The four phases differ primarily in the contemporary conceptions of the respective researchers regarding the best approach that could eventually lead to higher intelligence.
Furthermore, Figure \ref{1_timeline} shows a detailed timeline of the history of AI with a selection of different milestones and events.

The first phase in Figure \ref{1_ai_by_meth} Cybernetics (mainly between 1943 - 1955) can be seen as the pioneering stage before AI emerged.
Cybernetics is defined as the study of the fundamentals of the control and communication in animals and machines \cite{Wiener.1948}.
To this end, closed control loops are assumed where an agent acts on an environment to reach a goal based on the perceived feedback.
The overall goal in cybernetics was to create simple systems that mimic basic cognitive processes of brains.
Models from this epoch adopted ideas from a wide variety of different disciplines such as 
(i) (neuro-) biology, 
(ii) communication/information theory, 
(iii) game theory, 
(iv) mathematics - particularly statistics -, 
(v) philosophy - especially logic -, 
(vi) experimental psychology, and 
(vii) linguistics \cite{Buchanan.2005}.  
%
% TODO: - - - (\textit{think of using sources for every discipline}).
%
Due to the absence of computers, simple electronic networks were mostly used to examine and verify theoretical considerations.
An important milestone of this time (and an example for the adoption of biology) is the McCulloch and Pitts model that is today considered as the first Turing-complete digital model of neurons \cite{McCulloch.1943}.
A further popular representative of this epoch is SNARC, which is considered the first neural network computer \cite{Russell.2016}.
Although the cybernetics era was largely abandoned after 1956, some of the today's prevailing methods are significantly based on theoretical concepts of this phase. 
For instance, recently reached milestones in AI are based on deep reinforcement learning adopting both, neural networks and control loops (for instance, Atari \cite{Mnih.2015} or AlphaGo \cite{Silver.2016, Silver.2017}).

%expert systems that worked from datasets codifying human knowledge and practice to automate decision-making. 

Launchbury defined three waves of AI that occur since 1956, whereby the third wave is a projection of future developments that will be covered in the next section.
The first wave of AI (see Figure \ref{1_ai_by_meth}) is subject to handcrafted knowledge and mainly occurred after the \textit{Dartmouth Conference} (1956), a summer workshop that is considered being the founding event of AI \cite{JamesMoor.2006}.
In this epoch, the emergence of computers lead to the idea to express knowledge explicitly in the form of human-readable symbols.
In fact, the theory of symbol manipulation originally dates back to 1944, where it was originally formulated by Herb Simon (as mentioned by \textcite{Newell.1972}).
In this phase, agents emerged that were capable of solving a wide variety of tasks such as (i) logistics problems, (ii) theorem proving, (iii) or tax statement creation.
Two important milestones of this time are as follows:
(i) The\textit{ Logic Theorist} (1956) by \textcite{Newell.1956} is considered the first program that performed automatic reasoning.
This was demonstrated by proving mathematical theorems. 
A mathematical hypothesis serves as the starting point from which a tree is spanned whereas each path is built by the application of logical rules. 
If a goal was found - the respective path describes the prove. 
(ii) R1 (1978) by \textcite{McDermott.1980} is seen as the first example of a commercially exploited expert systems that was able to assist in the processing of new orders for computer systems.
%
% These systems were based on set of rules from which knowledge in the form of actions or instructions can be derived.
%
Systems based on hand-crafted knowledge are restricted to perceive information in a way they are programmed for and are limited regarding the ability to learn new knowledge or to abstract new findings from existing knowledge \cite{Launchbury.}.
It is worth noting that AI based on hand-crafted knowledge is by no means an abandoned field but is still a hot topic with milestones like IBM's Watson\footnote{Note that IBM's Watson also makes use of different methodologies such as statistical approaches but is mainly based on symbolic AI.} that competed human opponents in Jeopardy in 2011 (an introduced to IBM's Watson can be found in \cite{Ferrucci.2012}).
%
%Knowledge formalized in symbolic form makes it difficult to deal with uncertainty that is crucial to describe the real world.

The second wave of AI (see Figure \ref{1_ai_by_meth}) denotes the upcoming trend of statistical learning and mainly occurred after 1986.
Statistical learning is inherently different from systems based on hand-crafted knowledge as information is not stored in ontologies of symbolic data but as weights like the connection strength between neurons in neural networks.
Today's statistical models are exceptionally good at learning based on large data sets and can also generalize - to some degree - beyond previously seen examples by recognizing learned patterns.
Although the era of statistical learning has taken off after 1986, important theoretical work took place much earlier.
For example, a quite remarkable milestone was achieved by \textcite{Samuel.1959}, who wrote an agent that was able to play checkers at world class level.
In particular, the ability to learn was quite a remarkable feature at that time, as mentioned by \textcite{Buchanan.2005}.
The second important event was the introduction of the perceptron model (1958) by \textcite{Rosenblatt.1958} that describes in its basic form a single neuron and hence, can be seen as the corner stone of modern deep neural networks.
%
%It was demonstrated that basic image recognition could be solved employing a single perceptron. 
%
The perceptron model adopts major aspects of the aforementioned McCulloch and Walter Pitts model \cite{McCulloch.1943} and learns by applying the Hebbian learning rule (\textit{what fires together, wires together}) \cite{Hebb.1949}.
However, no learning scheme for multi-layered perceptrons (MLP) could be identified at that time and even the simple task of learning a XOR gate could not be achieved.
In 1986, the XOR problem was solved for the first time by MLPs with an invention called backpropagation.
Backpropagation exploits the chain rule to propagate the learning error through all layers of a deep neural network to infer learning gradients.
These learning gradients can then be used to perform gradient descent on the weights that are the learnable connections between neurons.
From that point on, the training of deep neural networks became possible, which led to numerous applications that have become an integral part of modern life.
About ten years after the introduction of backpropagation, support vector machines\footnote{Note, that various predecessor models of support vector machines were introduced much earlier.} were introduced by Cortes \& Vapnik \cite{Cortes.1995} and random forests by Tin Kam Ho \cite{Ho.1995}.
Both methods have had a great influence in the field of machine learning and are nowadays frequently exploited methods especially in the field of data science.
A general advantage of statistical methods compared to symbolic AI is the ability to express uncertainties that - among others - lead to a great progress in the field of perception.
For instance, \textcite{Launchbury.} stated that perception based on statistical learning have had a great impact on the DARPA challenge\footnote{The DARPA Grand Challenge was a competition for unmanned land vehicles sponsored by the Defense Advanced Research Projects Agency (DARPA) of the US Department of Defense.} whereas the following example was given:
The capability of discriminating between a black rock and a shadow was not straight forward by using conventional approaches but can be solved way more conveniently with statistical methods.
However, a disadvantage of statistical learning in comparison to symbolic-based methods is the incapability to reason.
For example, a neural network is not able to answer why an image is classified in showing a cat or a dog.
In this point, symbol-based systems are indeed advantageous as they are able to reason.
For instance, by predicting a blood infection using an expert system based on symbols\footnote{Mycin, invented by Feigenbau, Buchanan, and Shortliffe in 1973 \cite{Shortcliffe.1976} was an expert system that was able to diagnose blood infections.}
one can deduce the outcome by simply reviewing the decision process.
A further weakness of statistical learning is the inability to abstract as the underlying methods have no understanding of the matter.
Consider that the application of neural networks in computer vision is not based on the identification of features in an image.
To the contrary, humans efficiently classify pictures by recognizing the whiskers of cats or the tail of dogs.
%
%Another aspect is that statistical models can be sensitively fitted to the respective input data.
%%
%For instance, it can be shown that the performance of a neural network may catastrophically decrease by adding a small amount of noise (unrecognizable for humans) that is the most critical disturbance identified through reverse engineering of the neural network.

% Start with what is important today:
%
Another important impact was triggered by the adoption of scientific methods and best practices, particularly from the areas of mathematics and statistics \cite{Russell.2016}.
Publishing new research results in appropriate mathematical formalization is of great benefit because 
(i) existing methods from other disciplines (for instance, statistics, biology, neuroscience) can be more easily adopted, 
(ii) researchers can better understand complex concepts that are formulated in a common language, and 
(iii) it is easier to adopt/build on existing work.

Big data is a trending topic that emerged in the noughties and has been keeping the AI community on its toes ever since.
Large quantities of structured and unstructured data are generated primarily by the tremendous increase in devices and sensors connected to the Internet (IoT - Internet of Things), as well as by people feeding online services such as social media with their personal content.
With the help of big data, companies are able to make objective, data-based decisions without having to rely on intuition and subjective experience.
This enables faster and more efficient decision making, which can result in significant economic value \cite{OLeary.2013}.
Applications based on big data are becoming increasingly complex and require an ever-increasing amount of data.
Hence, data acquisition and automatic labeling techniques (for example, semi-supervised learning and active learning) became a critical factor for corporates \cite{Roh.2021}.
%

%
%In this context, algorithms that are published as open source are another new trend that is driving the field of AI.
%
Another influential trend is increasing openness.
Even leading commercial research teams are nowadays frequently publishing state-of-the-art methodologies.
Besides communication via scientific publications, the respective code is often published in the form of open source.
An example is given by OpenAI\texttrademark where openness is a central aspect of the mission statement.
But where is the motivation towards openness coming from? 
As mentioned by \textcite{Bostrom.2017}, openness has a positive impact on the overall reputation of the respective corporation, might lead to more attractive working conditions for researchers, and also lead to automatic benchmarking.
An obvious benefit of the emerging plethora of research papers and the corresponding source code is that researchers and companies are able to quickly become proficient in the latest methodologies.
For a more in-depth analysis of the impact of openness on AI, readers are directed to \textcite{Bostrom.2017}.

After reviewing historical and current movements with all their successes and glory, the following paragraph introduces a very essential aspect in the history of AI, namely the recurring AI winters.
The term AI winter refers to the hype cycles in which high expectations were built up that eventually remained unfulfilled.
During the hype phases, the fear of missing out initially prevailed, was accompanied by enormous private and public financial agendas, and eventually led to periods of depression.
A basic understanding of the reasons that lead to AI winters is important as the awareness could save us from repeating past mistakes or to be misled by ongoing hype phases.
However, AI, with its false narratives, coupled with the human psyche that seems to inherently crave hype, is a susceptible nexus that could be incorrigible in this regard.
Most references differentiate in two major AI winters.
The first AI winter took place mainly after 1969 and was mainly subject to three problems \cite{Russell.2016}:
(i) It was understood that systems based on symbolic knowledge, rules, and logic are incapable to understand the subject matter - or in other words to build a common sense.
In this context, the \textit{Chinese Room Argument} by \textcite{Searle.1980} states that a digital computer cannot gain any understanding of the challenge by simply following a syntax in a program.
Moreover, changing the syntax according to adapt to new challenges requires a deep understanding of the syntax itself.
A popular example of a natural language processing agent that fails to understand context is the famous translation of \textit{The spirit is willing but the flesh is weak} to \textit{the vodka is good but the meat is rotten}.
(ii) Computational complexity was not yet known and problems did not simply scale by employing
faster CPUs and more memory due to chain branching effects when the problem size increases.
(iii) As mentioned above, it was recognized that perceptrons - initially seen as universal function approximators - are not capable of solving a trivial XOR gate as already mentioned above.
The first AI winter culminated in a document by \textcite{Lighthill.1973} that was a commissioned report to evaluate the achievements of previous funding agendas.
The report concluded with a devastating assessment regarding the made progress in that time.
In particular, the combinatorial explosion of solving real-world problems is mentioned as a major reason to be pessimistic regarding further funding schemes.
As a result of the Lighthill report, funding schemes were suspended.
The second AI winter (around 1987) is associated with the formation of a bubble that formed after first successes of expert systems were presented and hundreds of companies with extravagant promises emerged. 
The winter started after it was realized that the capabilities of these expert systems were limited.
Many companies failed to deliver with what they have promised and an entire industry collapsed after stakeholders discontinued funding schemes \cite{Russell.2016}.

%\begin{wrapfigure}{r}{0.5\textwidth}
    %\centering%
    %\raisebox{0pt}[\dimexpr\height-1.5\baselineskip\relax]{\includegraphics[width=0.5\textwidth]{figures/1_school_of_thougts.pdf}}
    %\caption{The four school of thoughts by \textcite{Russell.2016}.}%
    %\label{1_school_of_thougts}%
%\end{wrapfigure}%
%
In the following, another way to distinguish AI systems based on the underlying motivation for creating the system is presented.
Four different schools of thought can be identified (adapted from \textcite{Russell.2016}):
(1) \textit{Thinking humanly}: In the thinking humanly approach, the human brain functions are focused.
To gather insights from the brain, different methodologies like determining own thoughts by introspection, psychological experiments, or by observing the brain in action (MRI, fMRI, EEG) are employed.
The overall idea and ultimate goal is to derive a theory of the mind with the purpose to create an artificial system that is capable of human thinking.
Note, researchers in this field do not intend to build systems that show optimal behavior in specific tasks.
Rather, researchers are aiming for a model similar to the imperfect human, in which the actual human thought process is mimicked regardless of the performance.
Truth to be told, brains are exceptionally complex and far from being understood and it is not yet possible to model/simulate a human brain.
However, many disciplines exploit results from cognitive science, such as neuro-physical findings that lead to developments in the field of computer vision \cite{Russell.2016}. 
(2) \textit{Acting humanly}: The creation of agents that behave like humans can be achieved without the necessity to rebuild what lead to this behavior and is hence different from the thinking humanly approach.
The agent's capability of acting humanly can be evaluated in the popular Turing test published in the seminal Paper by \textcite{Turing.1950} (originally coined as the \textit{imitation game}).
An agent passes the Turing test if an interrogator is not able to discriminate between an artificial agent that acts humanly and a real human.
For this purpose, an agent needs to master a variety of different disciplines like natural language processing, knowledge representation, automated reasoning (to make use of the stored knowledge), and machine learning (to adapt to specific situations by detecting patterns).
A further escalation of the standard Turing test is the total Turing test that includes visual interaction between the agent and the interrogator requiring the agent to employ computer vision and embodiment.
(3) \textit{Thinking rationally}: Deriving irrefutable conclusions from a knowledge representation is referred to as thinking rationally.
It is obvious that human thinking is irrational and hence, this approach is different from the thinking humanly approach.
In the context of AI, thinking rationally is often associated with using formal, symbolic representation of knowledge and to derive conclusions by applying logical rules.
An example for thinking in a rational way is to reason as follows \cite{Russell.2016}: \textit{Socrates is a man; all men are mortal; therefore, Socrates is mortal}.
As noted by \textcite{Russell.2016}, the main challenges of the so-called logicism approach are the representation of informal or uncertain knowledge.
So far, demonstration of basic principles is limited to trivial problems as real-world complexity requires a vast amount of data that is difficult to formalize in an automated way.
(4) \textit{Acting rationally}: Agents that act rationally do not mimic human thinking/acting but instead focus on acting in an optimal way towards an (expected) outcome/goal.
Note that acting rationally is more general than thinking rationally as the irrefutable inference is just one possibility to determine how to act rationally.
For instance, reinforcement learning can be shown to converge to optimal solutions without the ability to reason at all.
The acting rationally approach is more attractive for engineers building real-world applications than mimicking biological systems that involve complex and often unknown aspects.
An often-used analogy in this context (for instance in \cite{Russell.2016}) is the fact that aviation engineers initially studied how birds achieve flying.
Being inspired by having to wings did eventually not led to aircrafts that are exact mechanical copies of birds.
This analogy might also apply in the field of AI where agents are usually created to support humans in their daily lives rather than creating exact copies. 
It is worth mentioning that there are active communities in all four schools of thought, which can differ drastically in
terms of employed methods, assumptions, and opinions.

\section{Future Developments}
% Incorporate safety paper?

Projecting future developments in AI is an intriguing subject.
On the one hand, there is growing concern and fear about loss of privacy, autonomous military agents, or that humanity will become obsolete. 
On the other hand, disruptive new technologies could emerge that will change our daily lives for the better. 
A third scenario one can envision is that commercial exploitation of current methods is completed and a third AI winter ensues.
The future of AI is difficult to predict and an accurate projection cannot be given in this work.
However, an overview is provided of possible scenarios that might emerge to address the shortcomings of current methods.
In this context, Launchbury \cite{Launchbury.} predicts that the next era (the third wave of AI - see Figure \ref{1_ai_by_meth}) will be driven by models that are capable to understand context.
As mentioned above, statistical models cannot explain why an image is classified in showing a car or not.
Consider a model that is able to identify the most unique aspects of a car, such as four wheels, an antenna, multiple windows, et cetera.
Such a model would be substantially different from currently applied convolutional neural network (CNN).
A CNN starts by detecting edges, assembles these edges into more complex features such as circles and rectangles, reduces the information into several dense layers, and finally activates a single neuron that decides whether there is a car in the image or not.
While this approach is overly successful (an influential application can be found by \textcite{Krizhevsky.2017}), it requires an enormous amount of high-quality data that must be accurately labeled.
In contrast, a model that learns key aspects of the car only needs to understand what a tire is and how to recognize it. 
This differs from today's brute force methods where every scenario/view must be fed to the model (at least in a similar way as the target scenario) during training.
It is worth mentioning that a model identifying key aspects of an image is similar to the way human learn: 
Instead of pixel-wise feeding an image to the brain, human learn basic characteristics like recognizing the dot of the letter "i" \cite{Launchbury.}.
Moreover, such a methodology would address major shortfalls of current practices:
(i) 
Tremendous reduction of the required data as learning key aspects consumes less data than current deep learning methods.
The reasoning behind this is as follows:
A human learns the most important aspects of a car by seeing a few examples and identifying similarities. 
Intuitively, this is much more efficient than presenting every possible scenario/view in a training dataset.
(ii) 
Abandoning black-box models from which no explanation can be derived.
Consider that a model recognizing the whiskers of a cat can explain why a picture shows a cat rather than a dog.

Another often criticized aspect of today's AI applications is the limited range of problems that can be solved by a specific model.
As mentioned before, agents that can beat chess world champions tremendously fail in automating driving; vice versa, autonomous driving agents are usually very bad at playing chess.
In this context, a classification regarding AI methods that expresses the agent's capability to generalize on a wide variety of different problems can be made as follows:
(1) \textit{Artificial Narrow Intelligence} (ANI - also referred to as "weak" AI) comprises systems that are capable of solving specific tasks like playing video games, driving cars, recognition of speech et cetera.
Even though super-human performance can be achieved in various scenarios, ANI agents do not generalize on
experience to solve tasks that are different from the original task.
(2) \textit{Artificial General Intelligence} (AGI - also referred to as "strong" or human-level AI): AGI encompasses systems that are on the human level.
Agents within the AGI paradigm can solve new, unforeseen tasks by generalizing on their experience.
It is difficult to define what the attribute "human level" means.
Human level intelligence might be defined as the capability to autonomously set and pursue goals, reason on complex matters, to comprehend language, to perceive and make sense of the environment, to solve problems, or to make decisions under uncertainty.
One could require that an AGI agent must be integrated into the real world and mimic the sensorimotor capabilities of humans, for instance, to interact socially with other humans.
However, current methods cannot generalize, a fact that has recently led to a series of conferences on AGI \cite{Goertzel.2019}.
(3) \textit{Artificial Super Intelligence} (ASI):
ASI encompasses methods that go beyond human intelligence and is mentioned here only for the sake of completeness.
Discussions regarding ASI are often accompanied by complex philosophical implications.
Although ASI is frequently the subject of science fiction, it is also discussed on a scientific level for example by \textcite{Bostrom.2016} or \textcite{Tegmark.2017}.

Today's methods are in the field of ANI.
It has become the holy grail in AI to overcome this narrowness, while it is still unknown what could lead to higher generality.
Suppose for now that we follow the "human thinking" approach and are convinced that generality at the human level requires reverse engineering of the human brain.
The first question that arises is whether computers are inherently capable of performing the processes that occur in a brain.
Or, in other words, whether mental processes are substrate-independent and, if so, do not require a human brain \cite{Tegmark.2017}.
The \textit{Computational Theory of Mind}, formulated in a early version by \textcite{Putnam.1975}, raises the hypotheses that brains can be seen as information processing systems and can be mimicked by computers.
This theory is controversial and it is also argued that human reasoning is different from computation as it is not algorithmic \cite{Penrose.1989, Penrose.1994}.
For instance, \textcite{Weizenbaum.1976} pointed out that human thinking cannot be imitated by machines, since it relies on having some sort of wisdom to understand the whole picture and make the right decisions accordingly.
However, if the Computational Theory of Mind holds, we can conclude that any Turing complete system can represent a human-level intelligence given the right methodologies.
Consequently, the most obvious choice to reach human-level AI is to reverse engineer the brain.
However, a whole brain simulation is difficult to achieve mainly due to two reasons.
(i) The brain is exceedingly complex and not fully understood. 
Attempts to simulate the brain range from early approaches in cybernetics (for example by \textcite{McCulloch.1943}), over simplistic MLPs, to advanced non-stationary, bio-inspired approaches (such as spiking neural networks that account for synaptic plasticity).
(ii) Computational power is not yet sufficient to simulate the brain and it is not yet known how computational intense the undertaking is.
However, various estimates on the computational power necessary to emulate the brain exists (for instance by \textcite{Tegmark.2017, Russell.2016}).
Most of the projections imply that the computational resources will be available in this century.
It is also not clear whether an accurate computer model of a human brain will lead to something like consciousness.
Moreover, one might question which of the technologies might be the right path to higher intelligence.
Or to put it like this: Are more complex symbol manipulation systems or more data in statistical approaches are eventually leading to human-level intelligence?
%
%Nouvelle AI relies on the emergence of more global behavior from the interaction of smaller behavioral units.
%
An influential idea by \textcite{Brooks.1990} is that agents need to have an embodiment in order to interact with the real world by making use of the physical grounding.
It is argued that the grounding, for instance for symbols, is extraordinarily difficult to automated.
According to Brooks, grounding of symbols is not necessary as the physical representation of the environment is sufficient and the agent just needs to perceive it.
But does embodiment necessary lead to the emergence of human-level AI?
Let us consider a robot that learns to perform a task in a production line (for example based on reinforcement learning).
The agent has to learn the task from a large amount of experience that is sampled from the environment.
After the learning phase, the agent is looking up the action policy in a function that correlates status-action pairs with a predefined performance metric.
Although such an agent might be economically reasonable, it is important to realize that the agent is not building an understanding of the scenario and does not grasp which actions (causes) lead to which effects. 
The robot in the production line follows a specific syntax that fits a statistical model but can never come up with new suggestion for the program itself such as an improvement of the reinforcement learning algorithm.
In other words, it has no self-organizational capabilities and hence, cannot emerge.
We can conclude that although the robotic arm has an embodiment, no stronger intelligence emerges.
A further escalation of the embodiment theory is that a computer will never gain human-level intelligence as they are not part of the world like being integrated in a culture, having a childhood et cetera (\textcite{Fjelland.2020, JudeaPearl.2018, Dreyfus.1986}).

%Talk about cognitive systems...
%%
%Talk about Cognitive architectures....Readers interested in modern architectures aiming for generality are directed to \cite{TarekBesold.2016, Langley.2009}.
%%
%Read about enactive systems...

\section{Concluding Remarks}
This paper presents a bottom-up approach to introduce AI as an umbrella term that encompasses a plethora of different eras, narratives, methods, and schools of thought.
The key message of the first section is that the public portrayal of AI is flawed.
Products from big tech corporates changed our daily live and fueled a widespread misunderstanding of what "AI" comprises beyond applications in social media or e-commerce.
A critical view on public communication is recommended to the reader to maintain an objective view on current developments. 
In this context, two common arguments are examined: AI will lead to job losses and AI is emergent. 
Both arguments are often passionately debated, whereas the author believes that a more factual view would be appropriate.
The second section presents a brief history of AI.
It is evident that AI encompasses a variety of different aspects that tremendously complicate the understanding of the underlying terminology.
In this regard, the AI term is often reduced to the "acting rationally" school of thought, as this is the most visible aspect today.
%
%Moreover, the recurring pattern of AI winters is conveyed.
%%
%Undoubtedly, currently a substantial hype phase prevails.
%
Another addressed subject are AI winters that occurred regularly in the past.
From a historical point of view, it is questionable whether the current trend will continue or will lead to another phase of depression due to unfulfilled expectations.
On the one hand, there are clear shortcomings of the current trends, such as the lack of explainability, generalizability, or security mechanisms that could drive another AI winter.
On the other hand, it can be argued that the current hype cycle is different, as many business models have already proven to be commercially viable.
In the third section, the shortcomings of the current methods are highlighted in order to derive possible prospects for future advances. 
An important point is that universal methods that can generalize over tasks and domains are still far ahead.
Future advances in AI could be driven by the currently trending machine learning models when they will achieve higher data-efficiency, explainability, and context-awareness. 
Another approach would be to pursue bio-inspired methodologies, either through brain simulations (when higher computational power becomes available) or through hardware realizations (when neuromorphic hardware becomes accessible).
\printbibliography
\end{multicols}
\end{document}